%% file: 0_main.tex
\pdfoutput=1
\documentclass[10pt,twocolumn,letterpaper]{article}

\usepackage{pdfpages}

\usepackage{iccv}
\usepackage{times}
\usepackage{epsfig}
\usepackage{graphicx}
\usepackage{amsmath}
\usepackage{amssymb}
\usepackage[dvipsnames,table,xcdraw]{xcolor}
\usepackage{booktabs}
\usepackage{gensymb}
\usepackage{enumitem}

\usepackage{multirow}
\usepackage{tabularx,calc}
\usepackage{adjustbox}

\usepackage{booktabs,caption}
\usepackage[flushleft]{threeparttable}

\usepackage[pagebackref=true,breaklinks=true,letterpaper=true,colorlinks,bookmarks=false]{hyperref}

\newcommand{\calL}{\mathcal{L}}

\iccvfinalcopy 

\def\iccvPaperID{2641} 

\ificcvfinal\fi

\begin{document}

\title{MonteFloor: Extending MCTS for \\Reconstructing Accurate Large-Scale Floor Plans}

\author{Sinisa Stekovic$^{1}$, Mahdi Rad$^{1}$, Friedrich Fraundorfer$^{1}$, Vincent Lepetit$^{2,1}$ \\
$^{1}$Institute for Computer Graphics and Vision, Graz University of Technology, Graz, Austria \\
$^{2}$Universit\'e Paris-Est, \'Ecole des Ponts ParisTech, Paris, France \\
	{\tt\small \{sinisa.stekovic, rad, fraundorfer\}@icg.tugraz.at, vincent.lepetit@enpc.fr} \\
{\small Project page: \href{https://www.tugraz.at/index.php?id=52770}{ \color{blue} https://www.tugraz.at/index.php?id=52770}}
}

\maketitle
\ificcvfinal\thispagestyle{empty}\fi

\begin{abstract}
We propose a novel method for reconstructing floor plans from noisy 3D point clouds. Our main contribution is a principled approach that relies on the  Monte Carlo Tree Search~(MCTS) algorithm to maximize a suitable objective function efficiently despite the complexity of the problem. Like previous work, we first project the input point cloud to a top view to create a density map and extract room proposals from it. Our method selects and optimizes the polygonal shapes of these room proposals \emph{jointly} to fit the density map and outputs an accurate vectorized floor map even for large complex scenes. To do this, we adapt MCTS, an algorithm originally designed to learn to play games, to select the room proposals by maximizing an objective function combining the fitness with the density map as predicted by a deep network and regularizing terms on the room shapes. We also introduce a refinement step to MCTS that adjusts the shape of the room proposals. For this step, we propose a novel differentiable method for rendering the polygonal shapes of these proposals. We evaluate our method on the recent and challenging Structured3D and Floor-SP datasets and show a significant improvement over the state-of-the-art, without imposing any hard constraints nor assumptions on the floor plan configurations.
\end{abstract}

\input{1_Introduction}
\input{2_RelatedWork}
\input{3_Method}
\input{4_Evaluation}

\input{5_Conclusion}

{\small
\bibliographystyle{ieee_fullname}
\bibliography{string,egbib_cleaned}
}

\newpage~\newpage


\section*{Supplementary material}

# MonteFloor: Extending MCTS for Reconstructing Accurate Large-Scale Floor Plans
# –
# Supplementary Material

Sinisa Stekovic[1], Mahdi Rad[1], Friedrich Fraundorfer[1], Vincent Lepetit[2,1]
[1]Institute for Computer Graphics and Vision, Graz University of Technology, Graz, Austria
[2]Université Paris-Est, École des Ponts ParisTech, Paris, France
{sinisa.stekovic, rad, fraundorfer}@icg.tugraz.at, vincent.lepetit@enpc.fr

This document provides additional information about our MonteFloor method. We also provide a video demonstrating it in action.

## 1. MCTS Implementation

In this section, we provide more details about our implementation of Monte Carlo Tree Search (MCTS) with our refinement step.

**Tree nodes.** As explained in the main paper, the nodes contain several attributes. Every node is associated with some proposal $P_i$, but the same proposal $P_i$ is associated with one or more nodes along different sub-paths of the tree. Every node $N$ also keeps track of its "node score" $Q(N)$ and number of visits $n(N)$ that will be used later to guide the search towards the most promising directions. A node can have multiple children nodes children$(N)$ and has a single parent node.

**Initialization.** Initially, the search tree contains only a root node and is built online by expanding the tree in the most promising direction at every iteration.

During the search, we follow the standard Select-Expand-Simulate-Update strategy:

**Selection step.** At each iteration, and starting from the root node, we traverse existing nodes at each level of the tree by following the Upper Confidence Bound (UCB) criterion: From current node $N_{\text{curr}}$, we select its child node $N \in \text{children}(N_{\text{curr}})$ that maximizes the criterion

$$\arg\max_{N \in \text{children}(N_{\text{curr}})} Q(N) + \lambda_{UCB} \cdot \sqrt{\frac{\log n(N_{curr})}{n(N)}}, \quad (1)$$

where $\lambda_{UCB}$ is a hyperparameter that balances exploitation and the exploration during the search.

**Expansion step.** If during selection, there is a node with a child node $N \in \text{children}(N_{\text{curr}})$ that has not been created yet, we do not follow the selection criterion, but instead, we add $N$ to the tree. We set $Q(N) = -\infty$ and $n(N) = 0$ at this stage.

**Simulation step.** Immediately after the expansion step, we perform a simulation step and randomly create and traverse the children nodes until we reach a leaf node of the tree.

**Refinement step.** Whenever a leaf node is reached during simulation, we run 1 iteration of the Adam optimizer [2] on the selected proposals $P$ to minimize $\mathcal{L}(P)$ (Eq. (1) of the main paper) and calculate the score $S = -\mathcal{L}(P)$: Running more iterations decreases the number of necessary MCTS iterations, but still results in longer run-times. Instead, we allow the tree search to select already visited solutions. This results in faster run-time and accurate solutions, as we will run multiple refinement steps on the most promising solutions, but few on the less likely ones.

**Update step.** Once we know the score $S$ for a set of proposals $P$ after the refinement step, for every traversed node $N$, we update the node score $Q(N) \leftarrow \max(Q(N), S)$ and increment $n(N)$.

During experiments, for each scene we set $\lambda_{UCB} = 1$ initially, and linearly decay it such that the last MCTS iteration uses $\lambda_{UCB} = 0.01$.

## 2. Generating Floor Plans for Training the Metric Network

The metric network is trained on the Structured3D dataset [3] with input density maps and floor plans of size $256 \times 256$. During training, we generate input floor plans directly from the ground truth annotations to simulate a large variety of possible settings. More exactly, with probability of 30%, we select the ground truth floor plan as input. Otherwise, each individual room is added with probability of 50%. A single room is randomly rotated with 50%

chance by either $90°, 180°$, or $270°$, and randomly translated by $[-50, 50]$ pixels with probability of $10\%$. With $1\%$ probability, a vertex inside a room polygon is translated in range $[-10, 10]$ pixels. Labels for individual rooms are shuffled to make sure the network does not overfit to some specific label ordering. Similarly to input floor plans, we augment the dataset by rotating and translating the input density map and the corresponding ground truth floor plan. We match the generated room shapes with the ground truth shapes, as in Section 4.1 of the main paper, and calculate the Intersection-Over-Union (IOU) between the matches to obtain the final ground truth score.

The network is then trained by minimizing the Root-Mean-Square-Error (RMSE) between predicted and ground truth scores. We use the Adam optimizer [2] with learning rate set to $10^{-3}$.

## 3. Choice of Hyper-Parameters for the Objective Function

During refinement step, we set $\lambda_{\text{ang}} = 0.01, \lambda_0 = 0.01, \lambda_{\text{glob}} = 0.2, \lambda_f = 0.1$ and normalize $p(\alpha)$ to $[0, 1]$. We have found empirically that this set of hyper-parameters balances the influence of the corresponding loss terms. We have found that setting $\lambda_{\text{glob}} = 0$ and $\lambda_f = 1$ in the score calculation step increases convergence speed.

For the test set from [1], our metric network is less stable as it was not trained on this dataset. Hence, we have observed that setting $\lambda_{\text{ang}} = 0.1, \lambda_0 = 0.1, \lambda_{\text{glob}} = 0.2, \lambda_f = 0.05$, during refinement step, improves performance. During the score calculation step, we still set $\lambda_{\text{glob}} = 0$ and $\lambda_f = 1$.

## 4. Multiple Optimizers

As the number of room proposals increases, optimizer has to deal with much larger number of learning parameters. In turn, this leads to larger computation times during optimization step. To overcome this issue, we employ multiple optimizers: For each leaf node in the tree, we create an optimizer that optimizes the set of proposals along corresponding path in the tree. Such approach improves computation times considerably.

## 5. Qualitative Results

Figures 1 and 2 show additional qualitative results on the Structured3D dataset 1 and the test set from [1]. As shown in Figure 3, from the floor plans reconstructed by our MonteFloor method, we can directly reconstruct 3D floor plans simply by assuming constant wall height. Our 3D floor plans are consistent with underlying 3D scene.

## 6. Limitations and Future Work

We demonstrate some failure cases in Figure 4. Even though our method achieves state-of-the-art results on existing datasets we do believe there is still a large potential for improvements:

- The number of vertices of polygonal proposals is determined strictly by the polygonized outputs of the masks from Mask R-CNN. It would be interesting to consider possibility of dynamically adding, or removing, vertices of polygonal proposals. Similarly, it would be interesting to dynamically generate new rooms in places where large discrepancies are detected.

- It is clear from initialization that some proposals are unlikely to be part of the final layout. Using prior information about the quality of proposals to initialize node scores in the tree could make our approach run even faster.

- Density maps are a useful representation, but in practice, they can still be ambiguous in some cases. Hence, considering additional information such as perspective views, and camera trajectory, could be very helpful to deal with such ambiguities.

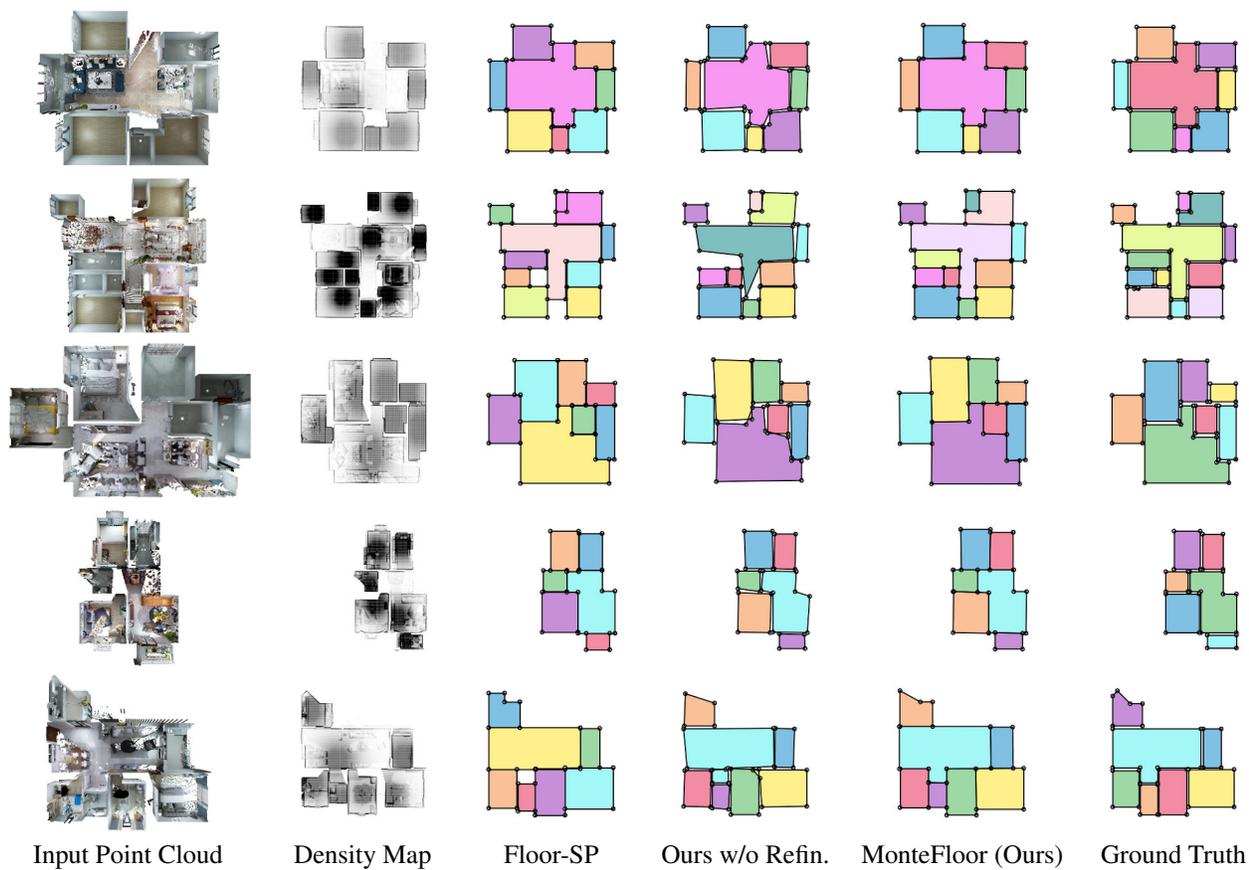

Figure 1: Qualitative results on Structured3D [3]. In the first example, both Floor-SP and our MonteFloor retrieve good floor plans. In the second example, Floor-SP misses some rooms and retrieves self-intersecting floor plans. In the third example, the light blue room reconstructed by Floor-SP that is slightly more complex than reality. In the fourth example, both methods perform well. In the fifth example, Floor-SP oversimplifies the yellow and dark blue rooms.

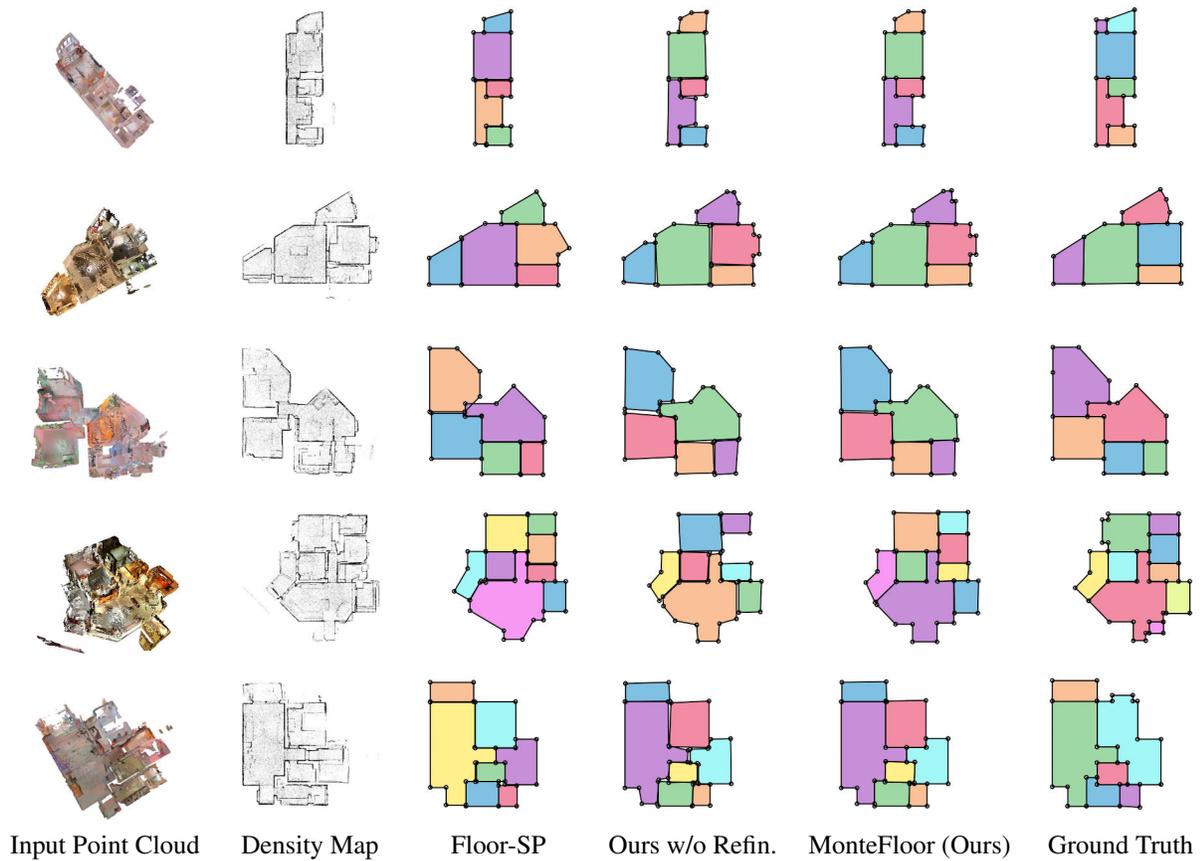

Figure 2: Qualitative results on the test set from [1]. In the first example, we believe that the floor plan retrieved by our method is better than the manual annotation as the purple room appears to be an annotation error and the light blue room is oversimplified. Similarly, in the second example, our reconstructions are much more consistent with the actual input. In the third example, Floor-SP produces self-intersecting rooms but our reconstructed green room is slightly more complex than its corresponding ground truth. In the fourth example, the little pink room on the bottom right of the ground truth is actually an annotation error. In the fifth example, the light blue room in the ground truth are actually two rooms, as estimated by both Floor-SP and our method. In addition, Floor-SP produces a small squared hole between the green, blue, red, and purple rooms.

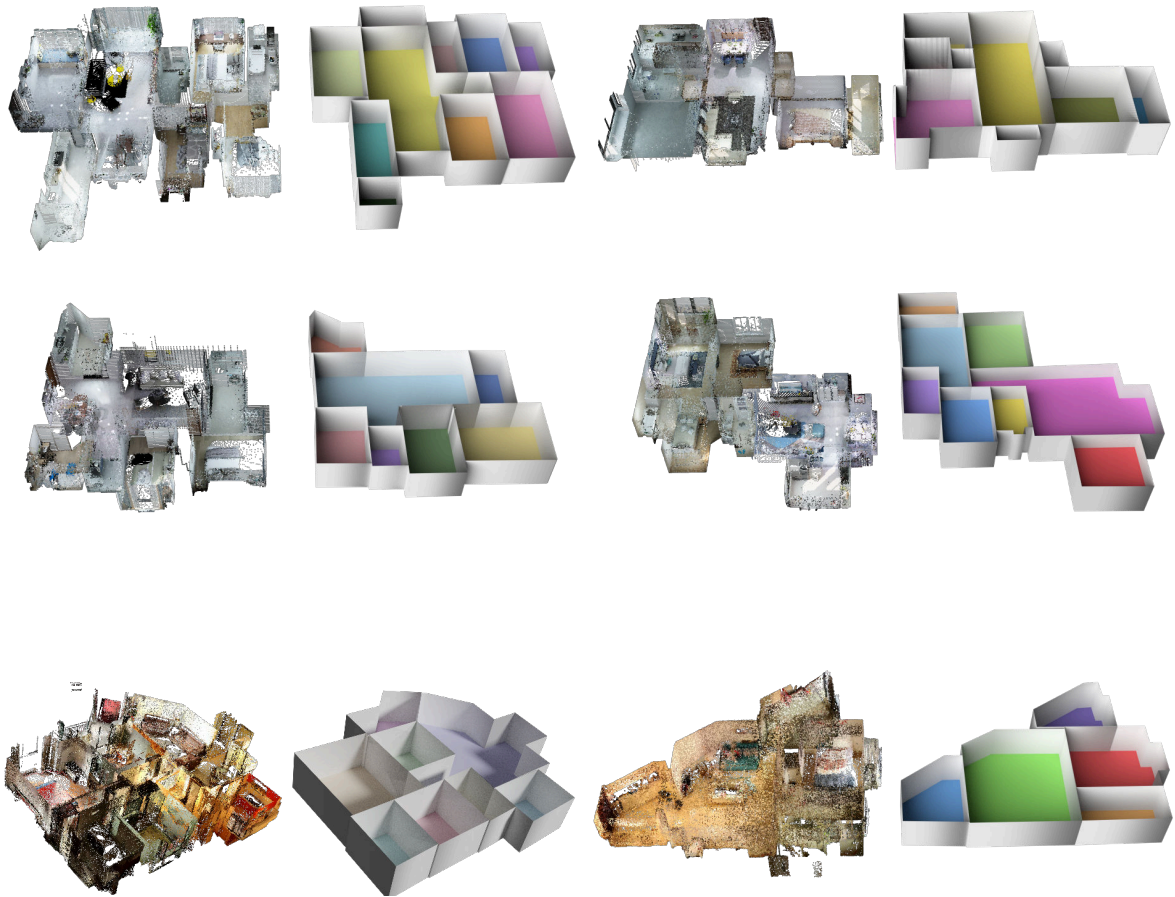

Figure 3: We can generate attractive 3D floor plans from our 2D reconstructions under a constant room height assumption. The examples demonstrate that the floor plans reconstructed by our MonteFloor method are indeed consistent with corresponding 3D scenes.

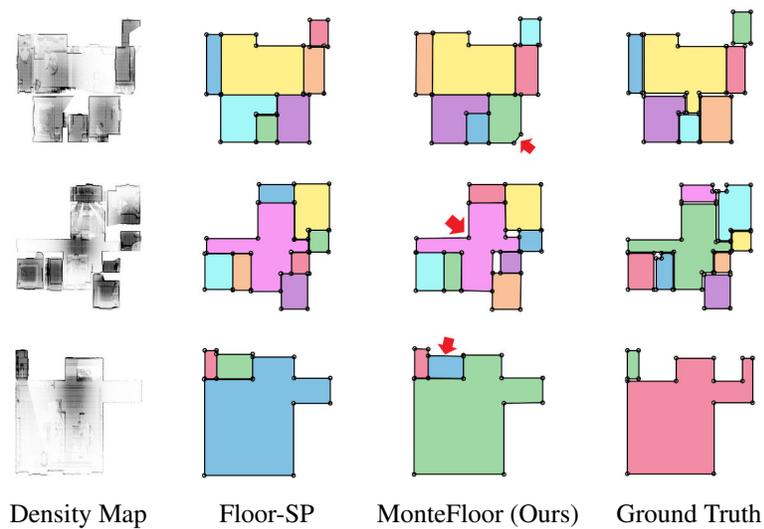

Density Map     Floor-SP     MonteFloor (Ours)     Ground Truth

Figure 4: Failure Cases. First row, red arrow: Our reconstruction of the green room is incorrect as there are some ambiguities in the density map for this region. In contrast, Floor-SP is more likely to produce Manhattan layouts in such situations. Second row, red arrow: None of the generated polygonal proposals for the pink room are adequate to represent the actual shape. Floor-SP performs graph-based search on pixel-locations that lie on the polygonal curve. As the final polygon is generated after the search, they obtain better reconstruction in this example. Third row, red arrow: Similarly to Floor-SP, we were not able to remove false positive proposal. This implies that the metric network could be further improved.



\end{document}

%% file: 1_Introduction.tex
\section{Introduction}
\label{sec:introduction}

Scene understanding from images is one of the main topics in computer vision, as it aims both at replicating one of the key abilities of human beings and producing solutions for many applications such as robotics or augmented reality. We focus here on the creation of a structured floor plan where each room of an indoor environment is represented as a polygon with one edge per wall. Many types of input have been considered: Monocular perspective color views~\cite{hedau09,Howard2018outsidethebox,Lee2017roomnet,stekovic_eccv20}, panoramic views~\cite{Sun2019horizonnet,Zou18,Zou20193d}, depth maps~\cite{Zhang2013estimating,Zou2019complete}. Here, we focus on unstructured 3D point clouds as in~\cite{ avetisyan2020scenecad,chen2019floor,Liu2018floornet,murali2017indoor}, as they can now be generated easily with an RGB-D camera and can cover an entire floor.

\begin{figure}
    \centering
    \begin{tabular}{ccc}
        \includegraphics[width=0.35\linewidth]{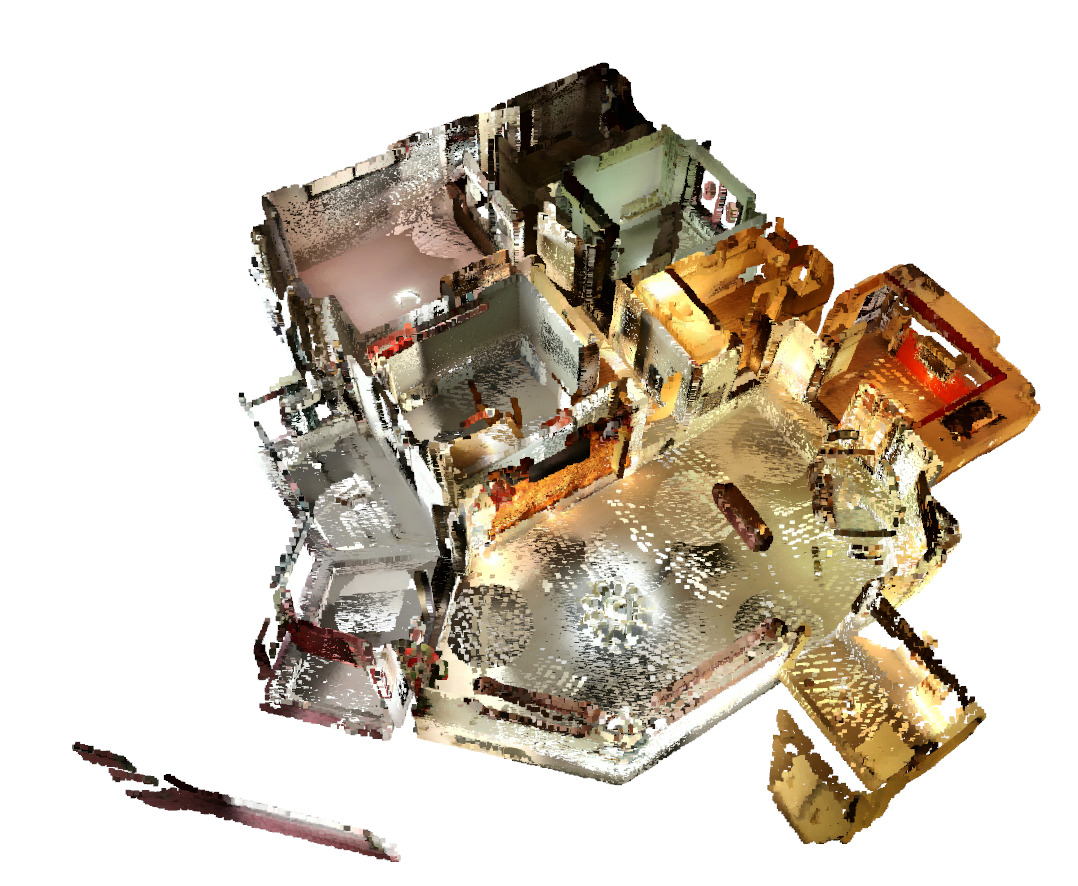} &  
        \includegraphics[width=0.25\linewidth]{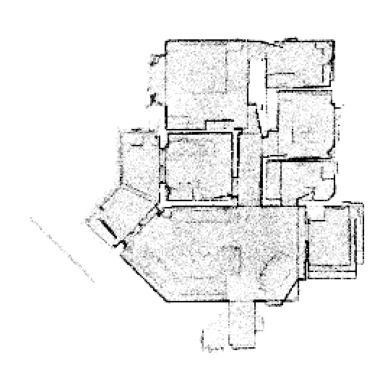} &
        \includegraphics[width=0.25\linewidth]{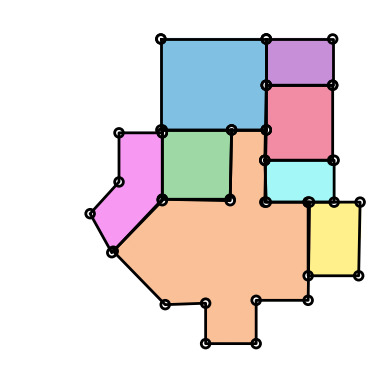} \\
        Point cloud & Density map & Our result \\
    \end{tabular}
    \vspace{-0.3cm}
    \caption{Given a density map \textit{i.e.} the top view of the 3D point cloud of a floor, we retrieve an accurate floor map that successfully recovers a variety of room shapes.}
    \label{fig:teaser}
\end{figure}

To estimate the floor plan from a given point cloud, \cite{chen2019floor,Liu2018floornet} proposes to first project the point cloud into a virtual top view to create a 'density map', as the walls, the main features for creating the floor map, appear relatively clearly in the density map. As shown in Figure~\ref{fig:teaser}, the density maps can be noisy, and it is still challenging to represent the rooms as vector drawings with a minimal number of edges as a human designer would do especially for non-Manhattan floor maps. To deal with this complexity, \cite{chen2019floor} proposes a graph-based solution with a sound energy term but still assumes the existence of some dominant wall directions in the scene.

In this paper, we also aim at estimating a floor plan from a density map. Our contribution is a method, which we call MonteFloor, that is conceptually simple and robust and  returns high-quality floor plans. Figure~\ref{fig:teaser} shows an example from the Floor-SP test set that demonstrates we can reconstruct complex floor maps, including very large ones with complex room shapes without having to tune hyperparameters.

Like \cite{chen2019floor}, our method starts from room proposals generated by Mask-RCNN~\cite{he2017mask} from the density map. However, the way we handle these room proposals is fundamentally different from \cite{chen2019floor}. Where \cite{chen2019floor} adjusts the room walls and corners in a greedy fashion, we select the correct room proposals \emph{jointly} \emph{while adjusting their locations and shapes}, guided by a learned scoring function. 

This is possible thanks to two main contributions. Our first contribution is based on the Monte Carlo Tree Search~(MCTS) algorithm~\cite{browne2012survey,Coulom06}. MCTS is a stochastic algorithm that efficiently explores search trees and has been used for example in AlphaGo and AlphaZero to select moves when playing Go or other games with high combinatorials~\cite{AlphaZero}. We use it to search among the room proposals the ones actually belonging to the correct floor plan. In our case, a move corresponds to the selection of a room proposal. In contrast with other tree search algorithms, MCTS is based only on the evaluations of leaves. This means that we can select a set of proposals based on how well they explain the density map \emph{together}.  After evaluating a leaf, MCTS updates a score in the visited nodes, which will be used to guide the next tree explorations. 

To evaluate how well a set of proposals explains an input density map, we introduce an objective function that combines a 'metric score' predicted by a deep network and regularization terms.  This network takes as input the density map and an image of the selected proposals to predict the Intersection-over-Union between the selected proposals and the ground truth. The regularization terms encourage the selected room proposals to be in contact with each other without overlapping, and angles close to 90$\degree$ to be exactly 90$\degree$---note this is different from enforcing Manhattan World conditions as other angles are also accepted.

Moreover, to adapt MCTS and obtain accurate plan estimates, we extended it by adding a refinement step before evaluating the objective function.  The step performs an optimization of the objective function and adjusts the shapes of the selected room proposals to better fit the density map. This is made possible by our second contribution, which is a novel differentiable method to optimize the shapes of 2D polygons.  Note that very recently, \cite{MCSS} has also used MCTS for a scene understanding problem. However, it proposes a straightforward application of MCTS. By contrast, we rely on a learned objective function suitable to our problem, and we introduce an optimization step to obtain accurate estimates.

While we focus in this work on floor plan estimation, we believe our approach is general and could be applied to other scene understanding problems, as its components are generic: We start from proposals for the target objects~(the rooms in this case). This step does not have to perform well to obtain good final results as our MCTS-based algorithm can deal with many false positives. This algorithm looks for the final solution by maximizing a data-driven score, which can thus be easily replaced to adapt to another problem. Our solution to refine the proposals is more specific to 2D polygons, but could inspire other authors to develop their own method adapted to their target objects.

To evaluate our method and compare it with Floor-SP~\cite{chen2019floor}, which is the state-of-the-art for our problem, we first perform experiments on the Structured3D dataset~\cite{Structured3D} that contains a variety of complex layout configurations. We show significant improvements  regarding both the accuracy and time complexity over Floor-SP (after retraining their method on Structured3D). As the authors of Floor-SP could not provide the training set for their method (as stated on their project page~\footnote{\url{https://github.com/woodfrog/floor-sp}}), we could not re-train our network for predicting the metric specifically for this dataset, and we had to use the one trained on Structured3D. Despite this domain gap, we achieve better performance on the Floor-SP test set without imposing any hard constraints nor assumptions on the floor plan configurations.

%% file: 2_RelatedWork.tex
\section{Related Work}
\label{sec:related_work}

Early methods for floor plan creation from 3D data relied on basic image processing methods such as histograms or plane fitting~\cite{ff:huber20113dreconstruction,budroni2010automated,ff:huber2009automatedmodeling,sanchez2012planar,xiao2014reconstructing,xiong2013automatic}. For example, \cite{ff:huber2009automatedmodeling} creates a floor plan by detecting vertical planes in a 3D point cloud by building a histogram of the vertical positions of all measured points. In a similar way, \cite{budroni2010automated} creates a floor plan from located walls in a 3D point cloud by extracting planar structures by applying sweeping techniques to identify Manhattan-World directions. However, these techniques relied heavily on heuristics and were prone to fail on noisy data.

Significant progress has been made later by using graphical models as in \cite{ff:cabral2014planarfloorplan,furukawa2009reconstructing,gao2016multi,gao2014jigsaw,ff:ikehata2015structuredindoor}.  \cite{furukawa2009reconstructing} uses graph-cuts optimization in a volumetric MRF formulation. However, the proposed method is vulnerable to noisy data, as regularization in MRFs is based only on pairwise interaction terms. \cite{ff:ikehata2015structuredindoor} combines an MRF with Robust Principal Component Analysis to obtain  more compact 3D models. 
Graphical models are also used in \cite{gao2014jigsaw} where layouts and floor plans are recovered from crowd-sourced image and location data. 

Graph-based methods define objective functions made of unary terms representing the elements of the plan and binary terms which involve only two elements at a time (here, the elements are mostly walls). In our case, we use MCTS as the optimization algorithm. MCTS does not impose restrictions on the form of the objective function and we use an objective function that captures complex constraints. In particular, the main term of our objective function is a deep network that considers all the elements at the same time. Moreover, we complement MCTS by adding a refinement step to adjust the locations of the elements based on the same objective function.

\input{3_1_overview_figure}

More recent works rely on other optimization techniques~\cite{ff:Chao2013layoutestimation,chen2019floor,Liu2018floornet}. The challenges for these techniques, however, are the definition of a cost function and the optimization procedure. One of these methods called FloorNet~\cite{Liu2018floornet} proposes a deep network for detecting probable corner locations from a given density map of the scene, followed by an Integer Programming formulation. However,  incorrect corner detection and misdetections result in missing or extra walls and rooms. Moreover, the solution space is restricted to Manhattan scenes and generalizing to non-Manhattan scenes would lead to a much larger solution space. By contrast, our approach is scalable, as it relies on the efficiency of MCTS to reduce the search space, and can consider Manhattan  and non-Manhattan scenes with the same complexity. It selects room detections that best explains the input through a global optimization, and is thus not sensitive to false positives.

The starting point of our method is inspired by Floor-SP~\cite{chen2019floor}, which proposes to first segment room instances, and then to reconstruct polygonal representations of rooms by sequentially solving shortest path problems. In their case, every pixel location in a discretized density map is a node in a graph that potentially belongs to the polygonal curve of the room. Wrong segmentations may still lead to an inaccurate floor plan structure, while we handle incorrect room segmentation at an early stage.  Also, Floor-SP discretizes the edge directions of rooms and models multiple Manhattan frames per room, while our approach can consider any angle. It still encourages angles close to 90$\degree$ to be exactly 90$\degree$, which results in better shapes when the rooms actually follow Manhattan structures while allowing other shapes. As we will show in the experiments, our approach outperforms the accuracy of Floor-SP.

\textbf{Differentiable Rendering. } Some works in 3D computer vision have shown interest in differentiable rendering~\cite{genova2020local,gkioxari2019mesh,groueix2018papier,mildenhall2020nerf,ravi2020pytorch3d,wang2018pixel2mesh}. However, these methods are focused on the rendering of 3D representations such as point clouds, voxels, meshes, and implicit 3D representations. In contrast, in this work we focus on fast differentiable rendering of 2D representations, \ie polygons, and introduce a differentiable winding algorithm for rasterization purposes.

%% file: 3_1_overview_figure.tex
\newlength{\nicemethodwidth}
\setlength{\nicemethodwidth}{0.25\linewidth}
\begin{figure*}
    \centering
    {\footnotesize
    \begin{tabular}{cccccc}
        \includegraphics[height=\nicemethodwidth]{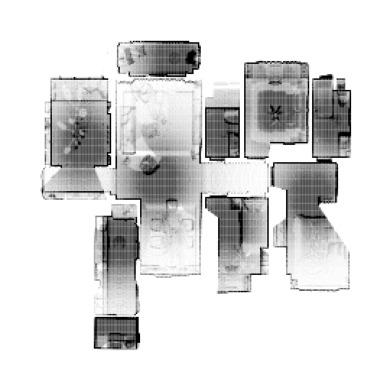} &
        \includegraphics[height=\nicemethodwidth]{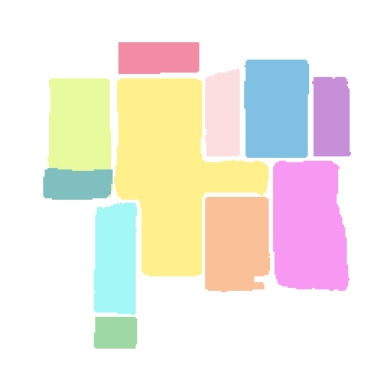} &
        \includegraphics[height=\nicemethodwidth]{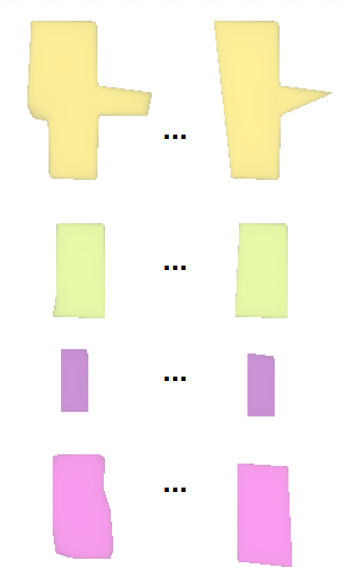} &
        \includegraphics[width=\nicemethodwidth]{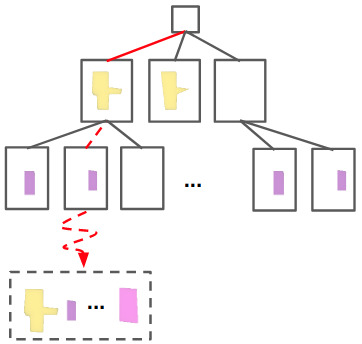} & 
        \includegraphics[height=\nicemethodwidth]{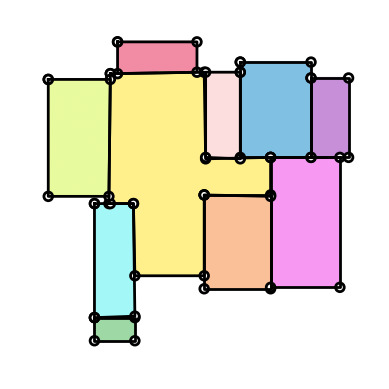} &
        \includegraphics[height=\nicemethodwidth]{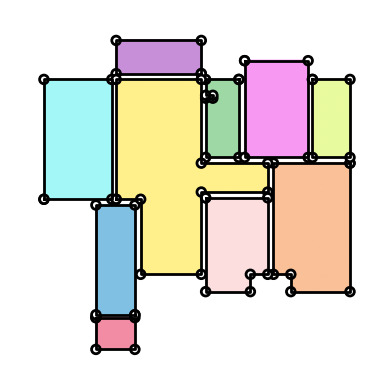} \\

        Density map & Detected & Some room proposals & Room selection with MCTS & Final result & Ground Truth\\[-0.05cm]
        & room segments &from polygonization  & + refinement & \\[-0.05cm]
        &&of the room segments&&\\[0.2cm]
    \end{tabular}
    }
    \caption{{\bf Overview of our MonteFloor method. } Given a 3D point cloud, we first create a density map of a floor. We then detect room segments using Mask-RCNN as in Floor-SP~\cite{chen2019floor}. Note the false positive at the bottom of the green segment on the left hand side. We polygonize each segment in different ways and obtain multiple room proposals from each room segment. We rely on MCTS and our objective function to select the correct room proposals, and our refinement step to adjust jointly the shapes of the room proposals to the input density map.}
    \label{fig:method}
\end{figure*}

%% file: 3_Method.tex
\section{Method}

\label{sec:approach}

\newcommand{\ang}{\text{ang}}
\newcommand{\reg}{\text{reg}}
\newcommand{\glob}{\text{glob}}
\newcommand{\TV}{\text{TV}}
\newcommand{\MSE}{\text{MSE}}
\newcommand{\calD}{\mathcal{D}}

\newlength{\progresstwidth}
\setlength{\progresstwidth}{0.10\linewidth}
\begin{figure*}
    \centering 
    \scalebox{0.85}{
    \begin{tabular}{ccc}
         \includegraphics[height=0.25\linewidth]{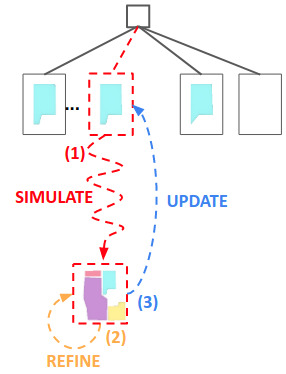} &  
         \includegraphics[height=0.25\linewidth]{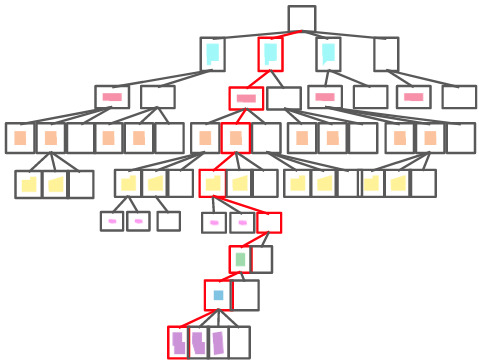} &
         \begin{tabular}[b]{ccccc}
         \multicolumn{5}{c}{\includegraphics[width=0.4\linewidth, trim=15 5 5 5, clip]{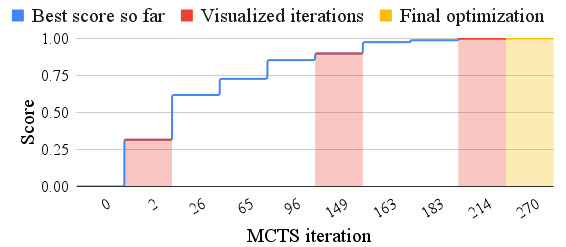}} \\
         \includegraphics[width=\progresstwidth]{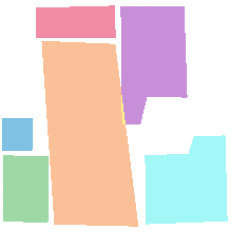} & 
         \includegraphics[width=\progresstwidth]{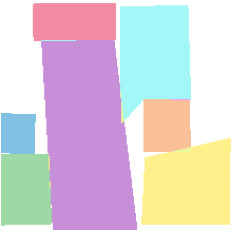} & 
         \includegraphics[width=\progresstwidth]{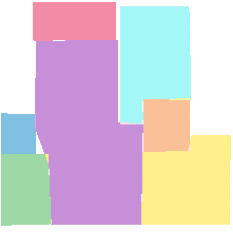} &
          \includegraphics[width=\progresstwidth]{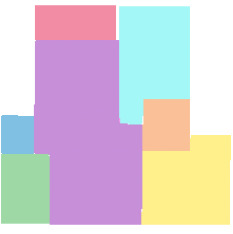} & 
         \includegraphics[width=\progresstwidth]{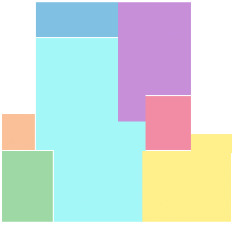} \\
         iter. 2 & iter. 149 & iter. 214 & final & g.t. \\
        \end{tabular} \\
         (a) & (b) & (c) \\
    \end{tabular}}
    \caption{{\bf Building the floor plan tree with MCTS}. (a) In our case, one node corresponds to the selection of a room proposal, or to skipping all the room proposals generated from a room segment. When a node is visited for the first time, MCTS runs a 'simulation' step. This step explores randomly the rest of the tree until reaching a leaf, in our case when there is no room proposal to consider any more. When reaching a leaf, we perform our 'refinement step', which optimizes the objective function over the room proposals in the path from the root node to the leaf. The value of the objective function is used to update the expected score for all the nodes in the path. (b) MCTS builds and explores only a portion of the tree. In contrast with other tree search algorithms, the pruning of MCTS is based only on the evaluations of leaves, which means that we can select a set of proposals based on how well they explain the density map together. (c) After few iterations, our MonteFloor method focuses at, and at the same time optimizes, solutions with promising expected scores. This enables us to quickly reconstruct an accurate floor plan of the scene, in about $60$ seconds for the scene used in this illustration.}
    \label{fig:tree}
\end{figure*}

Figure~\ref{fig:method} gives an overview of our MonteFloor method: Given a 3D point cloud of a scene, we first create a top-view density map of this point cloud, as explained later in Section~\ref{sec:density_maps}. We use Mask R-CNN~\cite{he2017mask} trained to detect rooms in such density maps and we polygonalize the detections  to obtain a set of room proposals. Some proposals will correspond, at least coarsely, to actual rooms but others are only false positives. We use MCTS to find which room proposals make together the best fit to the input density map. The MCTS search is guided by a 'metric network' trained to predict the Intersection-over-Union between the selected room proposals and the floor map ground truth. Because the shapes of the correct room proposals from Mask R-CNN correspond only coarsely to the real rooms, we optimise their shapes while performing the search in MCTS. This is done by introducing a differentiable method for rendering polygonal shapes.
 
In the following, we detail:
\begin{itemize}[topsep=0pt,itemsep=0pt,partopsep=0pt, parsep=0pt]
\item How exactly we obtain the room proposals;
\item How we use MCTS to select the room proposals;
\item Our objective function, involving our metric network and regularization terms;
\item How we refine the room proposals' locations and shapes within MCTS;
\item How exactly we compute the density map given a 3D point cloud.
\end{itemize}

\subsection{Generating the Room Proposals}
\label{sec:proposals}

We trained Mask R-CNN~\cite{he2017mask} on the density maps created from the training set of the Structured3D dataset~\cite{Structured3D} to extract individual room segments from a given density map. While resulting segments are of high quality, they can still contain false positives, however, they will be filtered by MCTS. Figure~\ref{fig:method} shows an example of room segments and the room proposals we generate from them. We detail this process below.

Sometimes, a room is detected as two segments that partially overlap. We thus merge two segments that overlap significantly (more than $5\%$ in practice) into an additional room segment, while keeping the two original segments.

In practice, the shapes of the true positive segments provided by Mask-RCNN do not correspond to the exact shapes of the rooms, as they are typically too smooth. We thus  polygonize the room segments to generate the room proposals. For this, we apply the Douglas-Peucker polygonalization algorithm~\cite{douglas1973algorithms} to the contours of the room segments. This algorithm depends on a parameter $\epsilon$ that controls the simplification of the contour. More exactly, this parameter is the maximum distance between the original curve and its approximation. As the exact complexity of the room shape is unknown at this stage, we generate multiple proposals from each segment by using different values for $\epsilon$. In practice, we take $\epsilon = d \cdot L$, where $d$ takes different values in a predefined set $\calD$ and $L$ is the perimeter of the segment, with $\calD = \{0.04, 0.02, 0.01\}$. Sometimes, 2 different $\epsilon$ result in the same number of vertices, and we keep only one of the two polygons.

Even after polygonalization, the shapes of the true positive room proposals may not correspond yet to the actual room shapes. To adjust their shapes, we will optimize them through our objective function. We describe the proposal selection by MCTS in the next subsection, and the objective function afterwards.

\subsection{Room Proposal Selection with MCTS}
\label{sec:joint}

MCTS is an algorithm to efficiently explore large trees where the score to maximize can be evaluated only for the leaves of the tree. We thus adapted it to select the room proposals based on an objective function used as the score. We describe this objective function in the next subsection.

As shown in Figure~\ref{fig:tree}, in our case, a move consists of selecting one of the room proposals generated by polygonizing one of the room segments. For each room segment, there is an additional move that consists of not selecting any of the room proposals from this segment.  The root node has up to $|\calD|+1$ children, corresponding to the  selection of one of the $|\calD|$ room proposals issued from the first room segment and the absence of selection from this room segment. 

The number of nodes of the full tree is at most $(|\calD|+ 1)^k$ where $k$ is the number of room segments, and, as $k$ increases, it quickly becomes infeasible to traverse all paths in the tree. Fortunately, MCTS will grow the tree only as needed while exploring it and avoids an exhaustive evaluation. We rely on the standard Select-Expand-Simulate-Update strategy, which we describe below briefly for completeness. For a more detailed description of the MCTS algorithm, we refer the interested reader to the survey in \cite{browne2012survey}. 

\paragraph{MCTS algorithm.} MCTS stores in the nodes the expected score of the patch they belong to, and uses them  for guidance towards an optimal selection.  As explained below, the expected score for new nodes is initialized using a simulation step, and can be updated after further exploration. At every iteration, starting from the root node, the tree is traversed  using the standard Upper Confidence Bound~(UCB) criterion to select each node. This criterion depends on the expected score stored in the nodes and balances exploitation and exploration. 

When reaching a new node, MCTS performs a simulation step to initialize the expected score for this node. This simulation step explores randomly the rest of the tree until reaching a leaf, in our case when there is no room proposal to consider anymore. We can then evaluate the score of the solution that contains the proposals selected in the path from the root node to the leaf. We explain in the next paragraph how we compute this score. The score is used to update the expected scores stored in the nodes of the path. We provide more details on our implementation of MCTS in the supplementary material.

\paragraph{Score and refinement step. } To compute the score of a solution corresponding to the path when reaching a leaf, we rely on our objective function that will be detailed in the next subsections. To obtain more accurate results, in addition to the standard MCTS steps, we introduce a refinement step that optimizes the objective function, before taking its value for the score of the solution: The locations and shapes of the room proposals may not correspond exactly to the actual rooms, and without this refinement it is possible that the value of the objective function is relatively low and not reflecting the actual quality of the selected proposals well. This refinement step adjusts the locations and shapes of the room proposals to obtain a more accurate solution.

\paragraph{Objective function.} Our objective function can be written as:
\begin{equation}
\calL(P) = - \lambda_f f(D, F(P)) + \calL_\text{reg}(P) \> ,
\label{eq:obj_fun}
\end{equation}
where $P$ is a set of room proposals for a solution to evaluate. $f(D, F(P))$ is our metric network, applied to the input density map and the floor plan of the room proposals $P$, weighted by $\lambda_f$. $\calL_\text{reg}(P)$ is a regularization loss. We detail both terms in the two next subsections. We use $-\calL(P)$ as the score maximized by MCTS.

\paragraph{Final solution inference.} After 500 MCTS iterations, we perform a final traversal through the tree following the nodes with the highest expected scores, and optimize the selected proposals by minimizing the objective function.
For some rare polygons, the vertices are less than $5$ pixels apart from each other. We merge the corresponding vertices to obtain the final solution.


\begin{figure}
    \centering
    \includegraphics[width=0.8\linewidth]{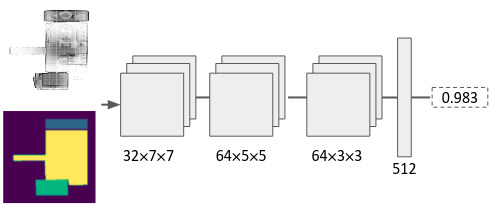}
    \caption{{\bf Our metric network.} This network takes a density map and a representation of the floor, colorized for visualized purposes, as input and outputs a score that measures how well the floor map fits the input density map. We train it to predict the Intersection-over-Union between the estimated floor plan and the ground truth.
    }
    \label{fig:scorenet}
\end{figure}

\subsection{Metric Network}

Our metric network $f(D, F(P))$ evaluates how well a set $P$ of selected room proposals fits the input density map. As shown in Figure~\ref{fig:scorenet}, this network has a simple architecture and takes two inputs: The first input $D$ is the density map. The second input $F(P)$ is an image of the room proposals, which we render  using their indices as pixel values:
\begin{equation}
F(P) = \sum_i i R(P_i) \> ,
\end{equation}
with $R(P_i)$ is a binary image of $P_i$, where the pixels inside $P_i$ are set to 1 and the others to 0. 

$f$ outputs only a single value, which should reflect the fitness between the room proposals and the density map. We train it to predict the Intersection-over-Union~(IOU) between the selected room proposals and the ground truth rooms for the density map in a supervised manner using training data from the Structured3D dataset~\cite{Structured3D}. More details on the training procedure can be found in the supplementary material. 


\subsection{Regularization Loss}

\newcommand{\ot}{\frac{1}{2}}
\newcommand{\of}{\frac{1}{4}}
\newcommand{\ps}{\frac{\pi}{6}}
\newcommand{\pt}{\frac{\pi}{2}}
\newcommand{\fpt}{\frac{5\pi}{12}}
\newcommand{\fps}{\frac{5\pi}{6}}
\newcommand{\sps}{\frac{7\pi}{6}}

The regularization loss $\calL_\text{reg}$ is decomposed into:
\begin{equation}
    \calL_\reg(P) = \lambda_\ang \calL_\ang(P) + \lambda_\glob \calL_\glob(P) + \lambda_0 \calL_0(P) \> ,
\end{equation}
where $\lambda_\ang$, $\lambda_\glob$, and $\lambda_0$ weight the three terms. We use the same weights for all the scenes and provide the actual values in the supplementary material.

$\calL_\ang(P)$ regularizes the angles of the room proposals in $P$:
\begin{equation}
    \calL_\ang(P) = -\frac{1}{|P|}\sum_{P_i \in P} \frac{1}{|P_i|} \sum_{(u,v,w) \in P_i} \log p(\widehat{(u,v,w)} ) \> ,
\end{equation}
where $|P_i|$ denotes the number of vertices in polygon $P_i$, $(u,v,w)$ denote any three consecutive vertices of polygon $P_i$, and $\widehat{(u,v,w)}$ their angle at vertex $v$. $p(\alpha)$ is a prior distribution we assume over the room angles. As shown in Figure~\ref{fig:reg}, we use a mixture of Gaussian distributions over their cosine and uniform distributions. It discourages flat angles (0$\degree$ and 180$\degree$), encourages right angles (90$\degree$ and 270$\degree$), and angles between $\pi/6$ and $5\pi/6$ and between $7\pi/6$ and $-\pi/6$ follow a uniform distribution. More formally, we take $p(\alpha) = $

{\normalsize
\begin{equation}
\frac{1}{Z}
\left\{
\begin{array}{ll}
G(\cos{\alpha} \;|\; \cos{\ps}, \sigma_1) & \text{if } \alpha \in ]-\ps;\ps] \>,\\ [0.1cm]
\eta + G(\cos{\alpha} \;|\; \cos{\pt}, \sigma_2) & \text{if } \alpha \in ]\ps;\fps]\>, \\[0.1cm]
G(\cos{\alpha} \;|\; \cos{\fps}, \sigma_1) & \text{if } \alpha \in ]\fps;\sps]\>, \text{ and} \\[0.1cm]
\eta + G(\cos{\alpha} \;|\; \cos{\-\pt}, \sigma_2) & \text{if } \alpha \in ]\sps;-\ps] \> ,\\[0.1cm]
\end{array}
\right.
\end{equation}
}
where $G$ denotes the Gaussian distribution, $\eta$ is the constant $G(\cos{\ps} \;|\; \cos{\ps}, \sigma_1)$, and $Z$ is a normalization factor. In practice, we use $\sigma_1=0.1$ and $\sigma_2=0.08$.

$\calL_\glob(P)$ encourages the room proposals to be in contact without overlapping. It can be seen in Figure~\ref{fig:reg} that the Total Variation~(the sum of the absolute values of the gradients) of an image of the proposals is a good criterion:
\begin{equation}
    \calL_\glob = \TV(F_1(P)) \> ,
\end{equation}
where $\TV$ denotes the total variation and $F_1(P)$ is an image of the proposals computed as 
\begin{equation}
F_1(P) = \sum_i R(P_i) \> .
\end{equation}
Figure~\ref{fig:reg} shows that this loss penalizes overlaps and pushes proposals toward each other, and by doing so, enforces similar orientations between the walls of neighbouring rooms.

\newlength{\tvwidth}
\setlength{\tvwidth}{0.2\linewidth}
\begin{figure}
    \centering
    {\footnotesize
    \begin{tabular}{@{}c@{}c@{}}
    \adjustbox{valign=c}{\includegraphics[width=0.35\linewidth,trim=30 5 50 10, clip]{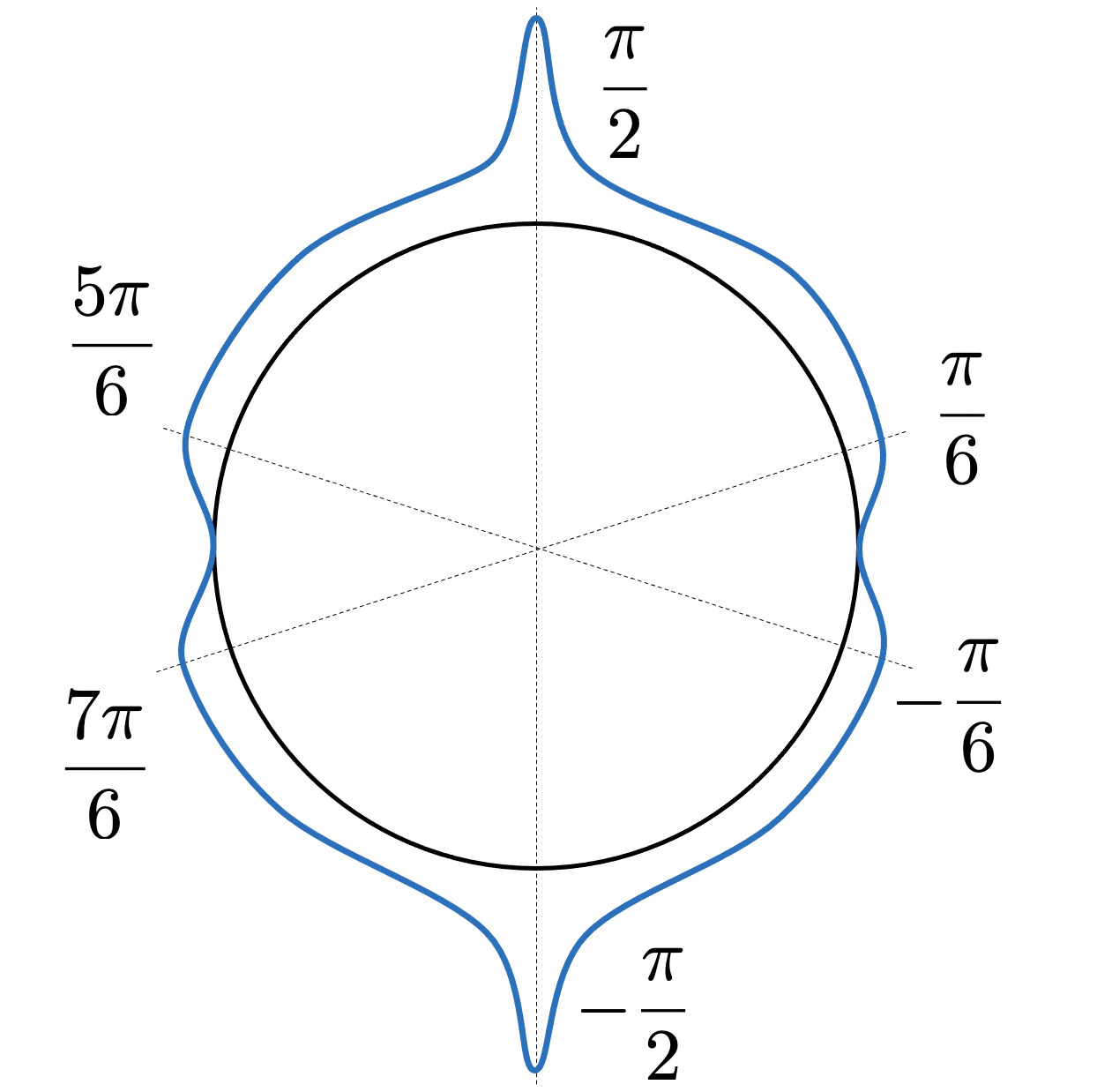}} &
    \begin{minipage}{0.6\linewidth}
        \begin{tabular}{@{}c@{}c@{}c@{}}
             \includegraphics[width=\tvwidth]{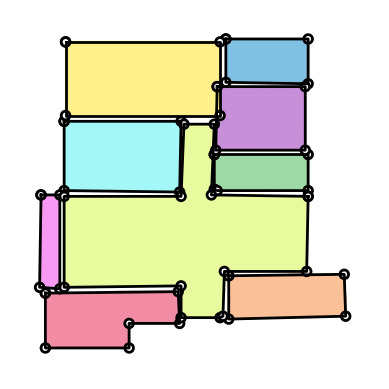}  &
             \includegraphics[width=\tvwidth]{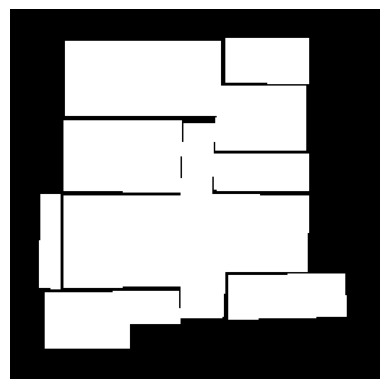}  &
             \includegraphics[width=\tvwidth]{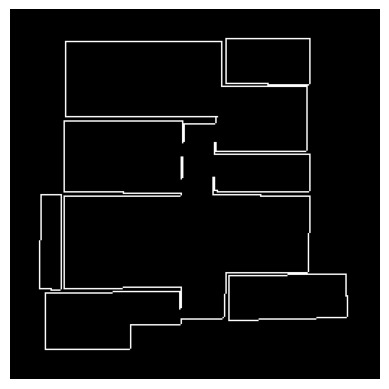}  \\
             
             \includegraphics[width=\tvwidth]{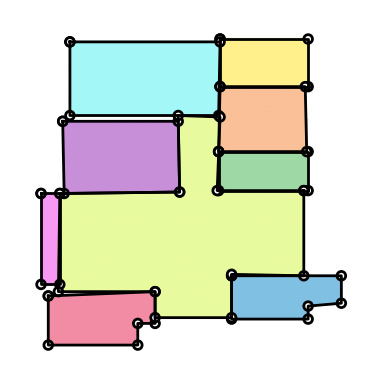}  &
             \includegraphics[width=\tvwidth]{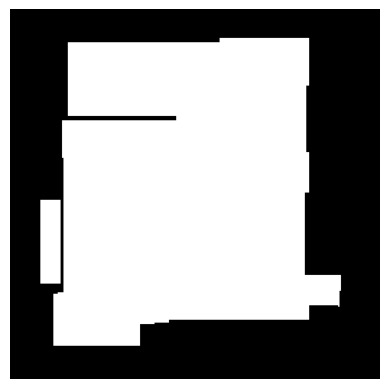}  &
             \includegraphics[width=\tvwidth]{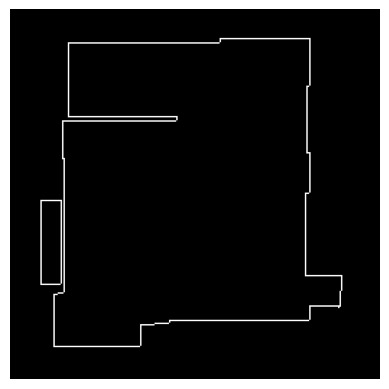}  \\

             $P$ & $F_1(P)$ & per pixel \\
                 &          & $\TV(F_1(P))$ \\
        \end{tabular}
    \end{minipage} \\
    (a) & (b)
    \end{tabular}
    }
    \caption{{\bf Visualization of the regularization losses.} (a) Prior distribution $p(\alpha)$ on angles discourages flat angles and encourages right angles, but other angles can still be accepted. (b) $\calL_\glob$ is based on total variation. {\bf Top:} When the room proposals in $P$ are not in contact or overlap, the Total Variation $\TV(F_1(P))$ of their image $F_1(P)$ is large. {\bf Bottom: } When the room proposals fit together, the Total Variation $\TV(F_1(P))$ is much lower.}
    \label{fig:reg}
\end{figure}

$\calL_0$ is used to prevent the proposals to drift from their initial locations during optimization. We take:
\begin{equation}
    \calL_0(P) = \frac{1}{|P|}\sum_{P_i \in P} \MSE(R(P_i), M_i) \> ,
\end{equation}
where $M_i$ is the binary image of the segment that generated proposal $P_i$, and $\MSE(\cdot)$ compares this image with the binary image $R(P_i)$ of the proposal.

\subsection{Refinement Step and Differentiable Polygon Rendering}

As explained earlier, when MCTS reaches a leaf, we perform several optimization steps of the objective function in Eq.~\eqref{eq:obj_fun} before computing its value and using it as a score for MCTS. In practice, we use the Adam optimizer~\cite{kingma2014adam}  for this task.

To optimize $\calL(P)$, we need to make it differentiable. The only part of it that is not trivially differentiable is the binary image creation $R(P_i)$ of a room proposal $P_i$, where $P_i$ is represented as a polygon. Differentiable rendering has already been developed~\cite{OpenDR}, however, available implementations are designed for rendering meshes of 3D triangles. Instead of tweaking these implementations to make them work on 2D polygons, we developed a much simpler approach by making the winding number algorithm differentiable. The original winding number algorithm checks whether a pixel location $m$ is inside a polygon $P_i$ by computing:
\begin{align}
    W(m, P_i) & = \frac{1}{2\pi}\sum_{(u,v) \in P_i} \text{sign}(\det(um, vm)) \widehat{(umv)} \> ,
\end{align}
where $(u,v)$ are any 2 consecutive vertices of $P_i$ and $\det(\cdot)$ is the determinant of vectors $um$ and $vm$. The $\text{sign}(\cdot)$ term is equal to 1 if angle $\widehat{(umv)}$ is between $]0;\pi]$ and to -1 if it is between $]\pi;2\pi[$, and $0$ otherwise. Hence, for a valid non-intersecting, closed, and counter-clockwise oriented polygon, $W(m, P_i) \in \{0,1\}$,  is a step function with value $1$ if $m$ is inside of $P_i$ and $0$ otherwise. 

To make it differentiable, we use the following expression instead:
\begin{align}
    W(m, P_i) & = \frac{1}{2\pi}\sum_{(u,v) \in P_i}
    \frac{c \cdot \det(um, vm)}{1 + |c \cdot \det(um, vm)|} \widehat{(umv)} \> .
\end{align}

The fraction term implements a soft form of the sign function that measures orientation of the triangle $(umv)$ with $c=1000$ to approximate the step form of the sign function in a differentiable way. To make rendering more efficient, we calculate the winding values only for pixel locations $m$ that are inside the bounding box detected by Mask R-CNN for the corresponding room segment.


\subsection{Computing a Density Map}
\label{sec:density_maps}

To obtain the density map $D$ of the scene, we follow a similar way to the one presented in~\cite{chen2019floor}. Given a registered set of RGB-D panorama images, we generate a point cloud of the scene. From the top-view of the scene's point cloud, we project the points to fit into a $256 \times 256$ image space, such that the top-view perspective remains unchanged and complete scene remains visible after the projection. The density value at a given pixel location is the number of points that projects to the same pixel location. The values of the density map are finally normalized to range $[0,1]$.

%% file: 4_Evaluation.tex
\section{Experiments}

In this section, we evaluate our method by comparing it to Floor-SP~\cite{chen2019floor}, the current state-of-the-art in floor plan reconstruction, on two datasets. We also provide an ablation study to show the importance of the refinement step for our method. 

\subsection{Metrics} 

To evaluate the quality of recovered floor plans, we first match the recovered rooms to the ground truth rooms. More exactly, starting with the largest ground truth room, we find the matching recovered room with the largest Intersection-Over-Union~(IOU) value. As we believe that the metrics used in~\cite{chen2019floor} are too permissive for really evaluating the quality of the compared approach, we made them more strict for quantitative evaluation:
\begin{enumerate}
    \item \textbf{Room metric.} This metric is the same as in \cite{chen2019floor}. A room polygon is considered to be successfully recovered if it is not overlapping other rooms and if it is matched with a ground truth room. We allow one pixel overlap between rooms and hence we do not penalize room polygons that are touching with this metric.
    
    \item \textbf{Corner metric.} A corner is considered to be successfully recovered if its corresponding room polygon is successfully recovered and it is the closest corner to any of the corners in the matching ground truth room polygon, within a distance of 10 pixels. This metric is inspired by the original metric from~\cite{chen2019floor} that did not consider if the corner actually belongs to the correct polygon.
    
    \item \textbf{Corner angle metric.} An angle of a room polygon is considered to be successfully recovered if its corresponding corner is successfully recovered and if the absolute difference to the corresponding ground truth angle is less than $5^\circ$.
\end{enumerate}

\input{4_quant_table}

\subsection{Evaluation and Comparison with Floor-SP}

\noindent \textbf{Structured3D.} We perform a first evaluation on the Structured3D dataset~\cite{Structured3D} that contains floor plan annotations for $3500$ scenes: $3000$ training scenes, $250$ validation scenes, and $250$ test scenes. To mimic the standard scene reconstruction pipeline, we project the registered RGB-D panorama images to obtain the point cloud of the scene. We process the reconstructions to obtain the training data for both Mask R-CNN and metric networks. For a fair comparison, we retrained the network used by Floor-SP for predicting the corner- and edge- likelihood maps on the training set generated from the Structured3D dataset and we replaced their Mask R-CNN network by ours also trained on Structured3D.

\noindent \textbf{Floor-SP test set.} Unfortunately, the authors of Floor-SP~\cite{chen2019floor} could not publish the training scenes for their Floor-SP dataset, but we could evaluate our approach on the $100$ publicly available test scenes, which include a large variety of floor plan configurations. We use the Mask R-CNN network pretrained on their training set as it was made available by the authors. However, since we could not train our metric network on the Floor-SP training set, we had to use the one trained only on Structured3D. Hence, the Floor-SP method has an advantage on this dataset.

\newlength{\niceresultwidth}
\setlength{\niceresultwidth}{0.18\linewidth}
\newcommand{\niceresult}[1]{
\includegraphics[width=\niceresultwidth]{figures/s3d_qual/#1_density.jpg} & 
\includegraphics[width=\niceresultwidth]{figures/s3d_qual/#1_fsp_edit.jpg} &
\includegraphics[width=\niceresultwidth]{figures/s3d_qual/#1_ours.jpg} &
\includegraphics[width=\niceresultwidth]{figures/s3d_qual/#1_gt.jpg} \\}

\newlength{\niceresultwidthfsp}
\setlength{\niceresultwidthfsp}{0.18\linewidth}
\newcommand{\niceresultfsp}[1]{
\includegraphics[width=\niceresultwidthfsp]{figures/fsp_qual/#1_density.jpg} & 
\includegraphics[width=\niceresultwidthfsp]{figures/fsp_qual/#1_fsp.jpg} &
\includegraphics[width=\niceresultwidthfsp]{figures/fsp_qual/#1_ours.jpg} &
\includegraphics[width=\niceresultwidthfsp]{figures/fsp_qual/#1_gt.jpg} \\}

\begin{figure}
    \centering
    \begin{tabular}{cccc}
    \niceresult{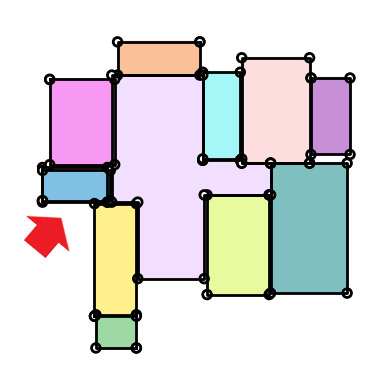}
    \niceresult{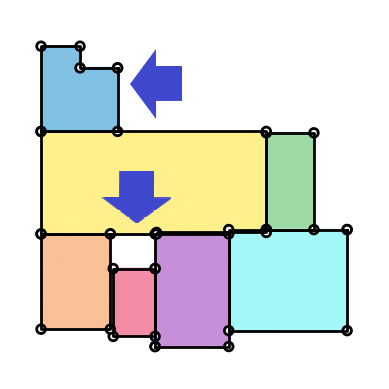}
    \niceresult{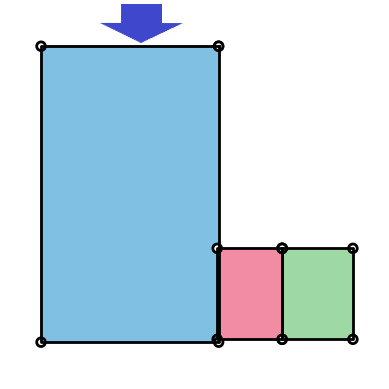}
    Input & Floor-SP & Ours & Ground Truth \\
    \end{tabular}
    \caption{\textbf{Qualitative results on the Structured3D dataset~\cite{Structured3D}, best seen in colour.} Red arrow: In contrast to Floor-SP, our approach deals well with false positive detections. Blue arrows: Compared to Floor-SP, we are able to model a larger variety of room shapes.
    }
    \label{fig:qual_sfp}
\end{figure}

\begin{figure}
    \centering
    \begin{tabular}{cccc}
    \niceresultfsp{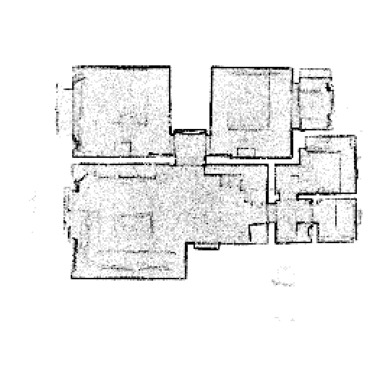}
    \niceresultfsp{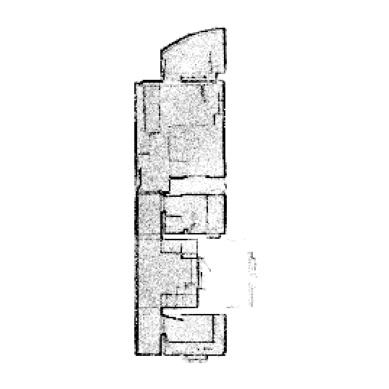}
    \niceresultfsp{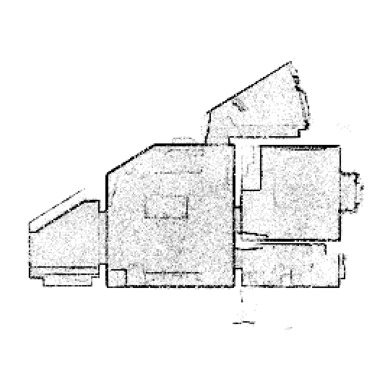}

    Input & Floor-SP & Ours & Ground Truth \\
    \end{tabular}
    \caption{\textbf{Qualitative results on the test set from~\cite{chen2019floor}}. Even though our metric network was not trained on the Floor-SP training set, our method still performs slightly better than Floor-SP on the Floor-SP test set. Red arrow: We remove false positive room detections. Green arrow: The purple room in the ground truth appears to be an annotation error. Blue arrows: Our reconstructions are sometimes more consistent with the input than the manually annotated rooms.}
    \label{fig:qual}
\end{figure}

Table~\ref{tab:quant} shows the quantitative results on both datasets. To better demonstrate the benefits of our approach, we also show the results of simple room detection by polygonization of room masks detected by Mask R-CNN with the Douglas-Peucker~(DP) approach that we used to initialize room proposals. DP obtains very high performance for the room metric, indicating that Mask R-CNN outputs masks of good quality most of the time. However, the angle metric clearly demonstrates that these polygons very often do not look anything like the actual room shapes. 

For Floor-SP, there is a drop in the room metric in comparison to the Douglas-Peucker method. This is related to the containment constraint satisfaction in the Floor-SP approach that forces the retrieved polygons to contain the segmentation mask completely. If this constraint cannot be enforced, the reconstruction will also fail. However, more importantly, the angle metric clearly demonstrates that their results are still superior to the ones obtained by DP.

Our approach outperforms both baseline methods by a large margin as we maintain very high performance on all metrics. This is true even for the Floor-SP test set, even though we could not retrain our metric network on the corresponding training set.

We improve performance on the room metric in comparison to Douglas-Peucker method as our refinement step adjusts the shapes of the room proposals that may initially overlap, and the selection by MCTS removes false positives. In contrast to Floor-SP, our approach benefits from optimizing directly on polygon shapes that enables us to avoid both the mask containment and angle discretization constraints. 

In addition, we compared the execution time of the two methods on the same machine. On the Structured3D dataset, the average computation time for Floor-SP is $785 \pm 549$ seconds. In contrast, the average computation time for our MonteFloor method is $71 \pm 40$ seconds, and $12 \pm 8$ seconds when skipping the refinement step. We made similar observations on the Floor-SP dataset.

\textbf{Qualitative results. } Figures~\ref{fig:qual_sfp} and \ref{fig:qual} show some qualitative results and  demonstrate that our approach is able to remove false positive detections and retrieve highly accurate polygonal reconstructions of floor plans.

\textbf{Ablation Study.} We performed an ablation study to evaluate the effectiveness of each individual term of our refinement procedure. As shown in Table~\ref{tab:ablation}, all of our regularization terms help to retrieve room polygons of better locations and shapes. The main ablation shows that the metric network has also a crucial role in our approach. Without the metric network, the objective function does not enforce consistency with the input scene. Then choosing a single correct room in a large scene maximizes precision as there are indeed no false positives, but minimizes recall. 

\newcommand{\without}{w\textbackslash o}

\begin{table}
    \centering
    \scalebox{0.7}
	     {
    \begin{tabular}{@{}l|c c c c c c | c c @{}}
    & \multicolumn{2}{c}{Room} & \multicolumn{2}{c}{Corner} & \multicolumn{2}{c}{Angle} & \multicolumn{2}{c}{MA} \\
    
    & Prec & Rec &  Prec & Rec & Prec & Rec & Prec & Rec \\
    \hline
    no refin. step   & 0.95  & 0.93  & 0.86  & 0.76  & 0.65  & 0.57  & 0.82  & 0.75  \\

    w/o $\calL_\ang$     & 0.96  & 0.94  & 0.86  & 0.75  & 0.73  & 0.68  & 0.85  & 0.79  \\

    w/o $\calL_\glob$   & 0.85  & 0.84  & 0.78  & 0.69  & 0.74  & 0.66  & 0.79  & 0.73  \\

    w/o $\calL_0$   & 0.92  & 0.92  & 0.87  & 0.76  & 0.84  & 0.72  & 0.88  & 0.80  \\
    
    w/o $f(.)$  & 0.94  & 0.22  & 0.89  & 0.15  & 0.87  & 0.15  & 0.90  & 0.17  \\

    complete & 0.96  & 0.94  & 0.89  & 0.77  & 0.86  & 0.75  & 0.90  & 0.82  \\

    \end{tabular}}
    \caption{\textbf{Ablation study.} Removing $\calL_\ang$ has a large influence on the angle metric;  Removing $\calL_\glob$ has a large influence on the locations of the corners; Removing $\calL_0$ may result in drift. Our metric network $f(\cdot)$ is crucial for the selection step of MCTS as the other terms are not a suitable scoring function for the floor plan generation task.
    }  
    \label{tab:ablation}
\end{table}

%% file: 4_quant_table.tex
\begin{table}[]
    \centering
    \scalebox{.7}
	     {
    \begin{tabular}{@{}l|c c c c c c | c c @{}}
    & \multicolumn{2}{c}{Room} & \multicolumn{2}{c}{Corner} & \multicolumn{2}{c|}{Angle} & \multicolumn{2}{c}{MA} \\
    & Prec & Rec &  Prec & Rec & Prec & Rec & Prec & Rec \\
    \hline
    \multicolumn{1}{l | }{{\bf Structured3D}}&\multicolumn{6}{l|}{}\\
    DP ($\epsilon = 0.01$) & 0.93  & 0.94  & 0.74  & 0.79  & 0.49  & 0.52  & 0.72  & 0.75  \\
    Floor-SP~\cite{chen2019floor}     & 0.89  & 0.88  & 0.81  & 0.73  & 0.80  & 0.72  & 0.83  & 0.78  \\

    MonteFloor (ours)  & \textbf{0.96}  & \textbf{0.94}  & \textbf{0.89}  & \textbf{0.77}  & \textbf{0.86}  & \textbf{0.75}  & \textbf{0.90}  & \textbf{0.82}  \\
    
    \hline
    \multicolumn{1}{l|}{{\bf \cite{chen2019floor} test set}}&\multicolumn{6}{l|}{}\\

    Floor-SP~\cite{chen2019floor}  & 0.85  & 0.83  & 0.72  & 0.58  & 0.65  & 0.52  & 0.74  & 0.64  \\

    MonteFloor (ours)  & \textbf{0.88}  & \textbf{0.85}  & \textbf{0.78}  & \textbf{0.63}  & \textbf{0.68}  & \textbf{0.54}  & \textbf{0.78}  & \textbf{0.67}  \\
    \end{tabular}}
    \caption{Quantitative results on Structured3D~\cite{Structured3D} and the test set from~\cite{chen2019floor}. MA is the average of the three metrics~(Room, Corner, and Angle). We compare our approach to a simple  Douglas-Peucker polygonization of the room segments obtained by Mask-RCNN~(DP) and to Floor-SP~\cite{chen2019floor}. Our approach slightly outperforms the other methods, even though we could not train our metric network on the training set of \cite{chen2019floor}. }
    \label{tab:quant}
\end{table}

%% file: 5_Conclusion.tex
\section{Conclusion} 

We proposed a method for floor plan estimation from 3D point clouds. We showed how we could apply the MCTS algorithm to this problem and how to add a refinement step to obtain accurate plans in a robust way.  

Beyond floor plan estimation, we believe our approach is general.
All that is needed to adapt it to other scene understanding problems is (1) a way to generate proposals and (2) a differentiable function to evaluate the quality of a solution.  We hope our work will inspire researchers to consider  problems with complex interactions between objects and obtain robust and accurate results. 

\textbf{\newline Acknowledgment.} This work was supported by the Christian Doppler Laboratory for Semantic 3D Computer Vision, funded in part by Qualcomm Inc.